\documentclass[11pt,a4paper]{article}
\usepackage[hyperref]{emnlp-ijcnlp-2019}
\usepackage{times}
\usepackage{latexsym}
\usepackage{subfigure}
\usepackage{graphicx}  
\usepackage{url}
\usepackage[export]{adjustbox}
\makeatletter
\newcommand{\ssymbol}[1]{^{\@fnsymbol{#1}}}
\makeatother

\aclfinalcopy 

\title{Machine Translation Evaluation Using Bi-directional Entailment}

\author{Rakesh Khobragade$^{\dagger}$, Heaven Patel$^{\dagger}$, Anand Namdev$^{\ddagger}$, Anish Mishra$^{\ast}$, Pushpak Bhattacharyya$^{\dagger}$ \\
  $^{\dagger}$ Center for Indian Language Technology (CFILT) \\
  Department of Computer Science and Engineering \\
IIT Bombay, India. \\
$^{\ddagger}$Microsoft, India.\\
$^{\ast}$SAP, India.\\
{\tt \{rkhobrag,heaven,anish,pb\}@cse.iitb.ac.in, annamdev@microsoft.com }
}

\date{}

\begin{document}
\maketitle

\begin{abstract}
In this paper, we propose a new metric for Machine Translation (MT) evaluation, based on bi-directional entailment. We show that machine generated translation can be evaluated by determining paraphrasing with a reference translation provided by a human translator. We hypothesize, and show through experiments, that paraphrasing can be detected by evaluating entailment relationship in the forward and backward direction. Unlike conventional metrics, like BLEU or METEOR, our approach uses deep learning to determine the semantic similarity between candidate and reference translation for generating scores rather than relying upon simple n-gram overlap. We use BERT's pre-trained implementation of transformer networks, fine-tuned on MNLI corpus, for natural language inferencing. We apply our evaluation metric on WMT'14 and WMT'17 dataset to evaluate systems participating in the translation task and find that our metric has a better correlation with the human annotated score compared to the other traditional metrics at system level. 
  
\end{abstract}

\section{Introduction and related work}
Automatic machine translation evaluation is highly useful for testing the quality of a translation system quickly, and hence for the overall development of the translation system. Many evaluation metrics have been proposed in the past, and all of them are compared against human judgement. A metric is deemed to be effective if it has a high correlation with the human judgement. 

Word error rate is one way to find normalized edit distance \cite{marzal1993computation} and is one of the earliest metrics applied for machine translation evaluation. It scores a candidate by finding words mismatch in the candidate and reference translation using Levenshtein distance \cite{levenshtein1966binary} divided by the length of the reference. Translation Error Rate (TER)\cite{snover2006study} is similar to word error rate, except that it considers a sequence of words instead of a single word. BLEU \cite{papineni2002bleu} is widely adapted metric for machine translation evaluation. BLEU generates an output score, matching n-grams in the reference and the machine translated sentence. This approach works well if there is a huge word overlap between the two sentences. Lexical similarity based metrics fail in situations where semantics are to be evaluated. Shortcomings of the BLEU metric have been put forward in various studies \cite{melamed2003precision, callison2006re, snover2006study, ananthakrishnan2007some}. METEOR \cite{banerjee2005meteor} addresses limitations of BLEU metric, to some extent but not completely, by a 3-stage matching process to match exact words, stemmed words and synonym words of the candidate and reference translation. LAYERED \cite{gautam2014layered} combines lexical, syntactic and semantic similarity into one metric using Universal Networking Language \footnote{http://www.undl.org/unlsys/unl/unl2005/UW.htm} (UNL). A neural network approach has been explored in \cite{guzman2017machine} to find better translation among a pair of candidates.

In this work, we propose a semantic similarity measuring technique and apply it for machine translation evaluation. We capture textual entailment between the candidate and reference sentences to find how close they are in meaning. Recognizing Textual Entailment (RTE) \cite{Dagan2006ThePR}, or natural language inferencing, is the task of determining if, for a given hypothesis and premise, the hypothesis can be inferred from the premise. Our approach uses deep learning to create sentence embedding for pair of sentences, which is then used for classification. We show that bi-directional entailment is an effective measure of semantic similarity and can be used for tasks like paraphrase detection and machine translation evaluation. 

\section{Bi-directional Entailment and Paraphrasing }
\begin{figure}%
\centering
\subfigure[]{%
\label{fig:a}%
\includegraphics[height=0.7in]{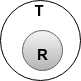}}%
\qquad
\subfigure[]{%
\label{fig:b}%
\includegraphics[height=0.7in]{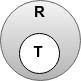}}%
\qquad
\subfigure[]{%
\label{fig:c}%
\includegraphics[height=0.7in]{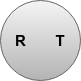}}%
\qquad
\subfigure[]{%
\label{fig:d}%
\includegraphics[height=0.7in]{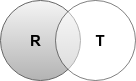}}%
\qquad
\subfigure[]{%
\label{fig:e}%
\includegraphics[height=0.7in]{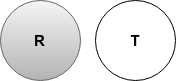}}%
\caption{Venn Diagrams explaining relationship between Reference (R) and Translation (T) sentences
\label{fig:1}}
\end{figure}

Entailment is a directed relationship between a pair of text. The entailed text captures a subset of information from the entailing text. As shown in Figure \ref{fig:1}, there can be different types of relationship between two text R (reference) and T (translation). These relationships indicate the overlap of information between two texts. Figure \ref{fig:1}.\ref{fig:c} refers to the case when both the texts are equivalent in meaning. It is the case of bi-directional entailment. Such equivalence can exist even when there is no or very less word overlap; for example, the following two sentences S1 and S2 are exactly the same in meaning but have very less word overlap.

\begin{center}
\textit{S1: Yoga is used to calm down body and mind}\\
    \textit{S2: Yoga helps you relax, physically and mentally}
\end{center}

Bi-directional entailment essentially determines if the two texts are paraphrases of each other since paraphrasing also require semantic equivalence. 
We propose a metric for machine translation evaluation, which finds equivalence between machine generated translation and reference translation. If reference and candidate sentences have bi-directional entailment between them, then they are semantically equivalent, which is a measure of adequacy.

\begin{table*}[h]
\centering
\scalebox{0.7}{
\begin{tabular}{|lllllll|}
\hline
\multicolumn{1}{|c|}{\textbf{Metric/Model}} & \multicolumn{1}{c}{\textbf{cs-en}} & \multicolumn{1}{c}{\textbf{de-en}} & \multicolumn{1}{c}{\textbf{fr-en}} & \multicolumn{1}{c}{\textbf{ru-en}} & \multicolumn{1}{c}{\textbf{Average}} & \multicolumn{1}{c|}{\textbf{SpearmanAvg}} \\ \hline
\multicolumn{1}{|l|}{BLEU} & 0.909 & 0.832 & 0.952 & 0.789 & 0.871 & 0.808 \\
\multicolumn{1}{|l|}{WER} & \textbf{0.992}$\ssymbol{2}$ & 0.762 & 0.952 & 0.809 & 0.879 & 0.775 \\
\multicolumn{1}{|l|}{TER} & 0.989$\ssymbol{8}$ & 0.772 & 0.951 & 0.810 & 0.880 & 0.779 \\
\multicolumn{1}{|l|}{PER} & 0.936 & 0.866 & 0.946 & 0.799 & 0.887 & 0.848 \\
\multicolumn{1}{|l|}{CDER} & 0.915 & 0.823 & 0.954 & 0.802 & 0.873 & 0.814 \\
\multicolumn{1}{|l|}{NIST} & 0.983 & 0.811 & 0.955 & 0.800 & 0.887 & 0.805 \\
\multicolumn{1}{|l|}{DiscoTK-Party-Tuned} & 0.975 & 0.943$\ssymbol{8}$ & 0.977$\ssymbol{8}$ & 0.870$\ssymbol{8}$ & 0.941$\ssymbol{8}$ & 0.902$\ssymbol{8}$ \\ 
\multicolumn{1}{|l|}{LAYERED} & 0.941 & 0.893 & 0.973 & 0.854 & 0.915 & 0.889 \\ 
\multicolumn{1}{|l|}{DiscoTK-Party} & 0.983 & 0.921 & 0.970& 0.856 & 0.933 & 0.875 \\
\multicolumn{1}{|l|}{\textbf{Bi-Di-Ent}} & 0.952 & \textbf{0.974}$\ssymbol{2}$ & \textbf{0.987}$\ssymbol{2}$ &  \textbf{0.912}$\ssymbol{2}$ &  \textbf{0.957}$\ssymbol{2}$ & \textbf{0.944}$\ssymbol{2}$ \\ \hline
\end{tabular}}
\caption{Pearson's System Correlation on WMT'14 data for baseline and participating metrics. Bold results with symbol $\ssymbol{2}$ denotes highest correlation and symbol $\ssymbol{8}$ denotes second highest correlation for a language pair.}
\label{table: system14}

\end{table*}

\begin{table*}[htbp]
\centering
    \scalebox{0.7}{
\begin{tabular}{|lllllllll|}
\hline
\multicolumn{1}{|c|}{\textbf{Metric/Model}} & \multicolumn{1}{c}{\textbf{cs-en}} & \multicolumn{1}{c}{\textbf{de-en}} & \multicolumn{1}{c}{\textbf{fi-en}} & \multicolumn{1}{c}{\textbf{lv-en}} & \multicolumn{1}{c}{\textbf{ru-en}} & \multicolumn{1}{c}{\textbf{tr-en}} &
\multicolumn{1}{c}{\textbf{zh-en}} &
\multicolumn{1}{c|}{\textbf{Average}} \\ 
\hline
\multicolumn{1}{|l|}{BLEU} & 0.971 & 0.923 & 0.903 & 0.979 & 0.912 & 0.976 & 0.864 & 0.932 \\
\multicolumn{1}{|l|}{AutoDA} & 0.438 & 0.959 & 0.925 & 0.973 & 0.907 & 0.916 & 0.734 & 0.836 \\
\multicolumn{1}{|l|}{BEER} & 0.972 & 0.960 & 0.955 & 0.978 & 0.936 & 0.972 & 0.902$\ssymbol{8}$ & 0.953 \\
\multicolumn{1}{|l|}{Blend} & 0.968 & 0.976$\ssymbol{8}$ & 0.958$\ssymbol{8}$ & 0.979 & 0.964$\ssymbol{8}$ & 0.984$\ssymbol{8}$ & 0.894 & \textbf{0.960}$\ssymbol{2}$ \\
\multicolumn{1}{|l|}{bleu2vec\_sep} & 0.989 & 0.936 & 0.888 & 0.966 & 0.907 & 0.961 & 0.886 & 0.933 \\
\multicolumn{1}{|l|}{CDER} & 0.989 & 0.930 & 0.927 & 0.985 & 0.922 & 0.973 & \textbf{0.904}$\ssymbol{2}$ & 0.947 \\
\multicolumn{1}{|l|}{CharacTER} & 0.972 & 0.974 & 0.946 & 0.932 & 0.958 & 0.949 & 0.799 & 0.932 \\
\multicolumn{1}{|l|}{chrF} & 0.939 & 0.968 & 0.938 & 0.968 & 0.952 & 0.944 & 0.859 & 0.938 \\
\multicolumn{1}{|l|}{chrF++} & 0.940 & 0.965 & 0.927 & 0.973 & 0.945 & 0.960 & 0.880 & 0.941 \\
\multicolumn{1}{|l|}{MEANT\_2.0} & 0.926 & 0.950 & 0.941 & 0.970 & 0.962 & 0.932 & 0.838 & 0.931 \\
\multicolumn{1}{|l|}{MEANT\_2.0-nosrl} & 0.902 & 0.936 & 0.933 & 0.963 & 0.960 & 0.896 & 0.800 & 0.912 \\
\multicolumn{1}{|l|}{ngram2vec} & 0.984 & 0.935 & 0.890 & 0.963 & 0.907 & 0.955 & 0.880 & 0.930 \\
\multicolumn{1}{|l|}{PER} & 0.968 & 0.951 & 0.896 & 0.962 & 0.911 & 0.932 & 0.877 & 0.928 \\
\multicolumn{1}{|l|}{TER} & 0.989 & 0.906 & 0.952 & 0.971 & 0.912 & 0.954 & 0.847 & 0.933 \\
\multicolumn{1}{|l|}{TreeAggreg} & 0.983 & 0.920 & \textbf{0.977}$\ssymbol{2}$ & 0.986$\ssymbol{8}$ & 0.918 & \textbf{0.987}$\ssymbol{2}$ & 0.861 & 0.947 \\
\multicolumn{1}{|l|}{UHH\_TSKM} & \textbf{0.996}$\ssymbol{2}$ & 0.937 & 0.921 & \textbf{0.990}$\ssymbol{2}$ & 0.914 & \textbf{0.987}$\ssymbol{2}$ & 0.902$\ssymbol{8}$ & 0.949 \\
\multicolumn{1}{|l|}{\textbf{Bi-Di-Ent}} & 0.994$\ssymbol{8}$ & \textbf{0.979}$\ssymbol{2}$ & 0.953 & 0.955 & \textbf{0.990}$\ssymbol{2}$ & 0.932 & 0.893 & 0.956$\ssymbol{8}$ \\
\hline
\end{tabular}}
\caption{Pearson's System Correlation on WMT'17 data for baseline and participating metrics. Bold results with symbol $\ssymbol{2}$ denotes highest correlation and symbol $\ssymbol{8}$ denotes second highest correlation for a language pair.}
\label{table: system17}
\end{table*}

\section{Approach}
In WMT metric task, translations from systems participating in translation task are provided for evaluation, along with reference translations. Our metric scores these translations by finding their similarity with the given reference translations. Human evaluation score is also provided for a subset of translations. The effectiveness of our metric is judged based on its correlation with human evaluation. Pearson's correlation \cite{pearson1931test} is used to compare different metrics. Our approach differs from \cite{guzman2017machine} in that we do not need a pair of translation candidates as input and our score is for the overall system.

We score candidate translation by determining if it is a paraphrase of the reference translation, such that correct paraphrasing receives a higher score. To do this, we check if the entailment relationship holds between the candidate and a reference translation in the forward and backward direction. We fine-tune BERT's \cite{devlin2018bert} pre-trained model for sentence pair classification on Multi-Genre Natural Language Inference (MNLI) \cite{mnli} corpus. BERT uses transformer networks \cite{vaswani2017attention} to generate a sentence embedding using multi-head attention. On MNLI corpus, this gives a test accuracy of $83.4\%$.

\begin{figure}
    \centering
    \includegraphics[scale=0.30]{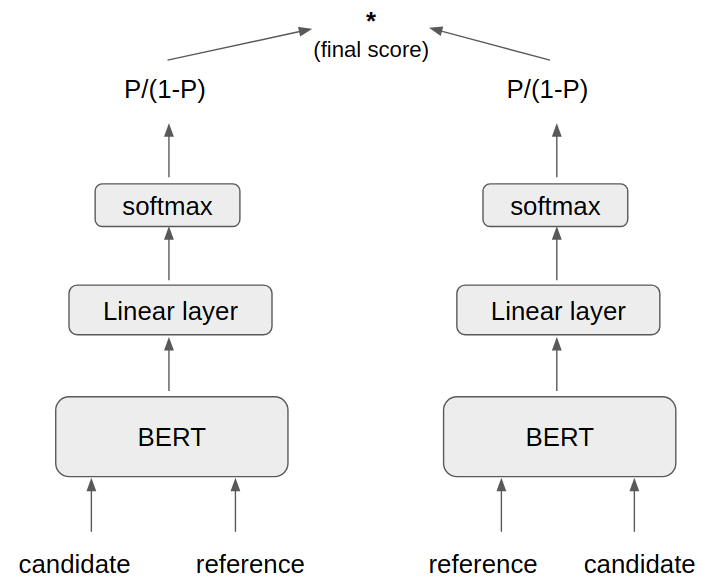}
    \caption{Score generation from a candidate and reference. P represents probability of entailment}
    \label{fig:blck}
\end{figure}

Figure \ref{fig:blck} shows the architecture of our system. The linear layer has 3 output neurons for labels \textit{contradiction, entailment} and \textit{neutral}. We apply softmax on these outputs to get the probability of entailment label and calculate the odds of candidate entailing the reference. We perform the same operation again, but we switch the candidate with reference to get odds of entailment in backward direction. It gives us odds of reference entailing the candidate. Product of these two odds is normalized across all examples to get the final score.
$$Score = odds_f * odds_b$$
Where, $odds_f$ is the odds of entailment in the forward direction and $odds_b$ in the backward direction. The odds are calculated from the entailment probability $P$, obtained from the softmax layer, as $odds = \frac{P}{1-P}$. Odds value is always greater than the probability value and it allows us to assign higher scores to candidates in which entailment probability is high compared to candidates that do not have entailment relationship. Average of scores across all test examples gives the overall system score. To judge our metric, this score is correlated with the score given by human to the system.

\section{Experiments and Results}
In our experiments, we use MNLI corpus for training and WMT(2014 and 2017) metric task data for testing. MNLI corpus consists of 433k annotated sentence pairs called premise and hypothesis and categorizes relation between them into three types: \textit{entailment, contradiction} and \textit{neutral}. Unlike its predecessor SNLI \cite{bowman2015large}, MNLI contains sentences from multiple genre. WMT dataset contains the output of machine translation systems for different language pairs and human translated references. WMT'14 \cite{machacek2014results} includes five such language pairs in which we exclude ``hi-en'', whereas WMT'17 \cite{wmt17results} dataset includes seven language pairs. Our metric evaluates translations to English language. Translation evaluation metrics are judged at two levels: segment and system. Our work focuses on system level scoring.

\begin{table*}[h]
\begin{tabular}{p{5cm}p{5cm}p{1.8cm}p{1.8cm}}
\hline
Reference & Candidate & Human evaluation & Bi-Di-Ent score \\ \hline
Mehmet Şimşek: Turkey will not break loose from the West, will not give up on the European Union & Mehmet Simsek: Turkey does not break from the West, does not give up on the European Union & 0.913 & 4.559 \\ \hline
Phelps' historic achievement could not be overshadowed even by his compatriot Ryan Murphy, who won the 100m and 200m backstroke titles in Rio. & Legendu could not shadow even compatriot Ryan Murphy, who won the doubles double - after a hundred won in Rio also a double track. & -1.586 & 3.462e-09 \\ \hline
The new location has other perks. & There are other advantages to the new location. & 1.190 & 12.639 \\ \hline
\end{tabular}
\caption{Human evaluation score and Bi-Directional Entailment(Bi-Di-Ent) score for examples from WMT'17. Human evaluation score is averaged across multiple human scores.}
\label{table: examples}
\end{table*}

\subsection{Training Details}
We use BERT's bert-base-uncased pre-trained model and fine-tune it on MNLI corpus. The learning rate is 2e-5, maximum sequence length is 128, and the batch size is 8. Number of epochs is 3. BERT's tokenizer is used to perform end-to-end tokenization which involves basic tokenization followed by word tokenization.

\subsection{Results}
The results for WMT'14 is shown in Table \ref{table: system14}. Our metric shows better correlation with the human score than all other evaluation metrics for all language pairs except for cs-en pair. Our metric has an average correlation of 0.957, which is an improvement from 0.871 of BLEU. Compared to LAYERED, which uses UNL to capture semantics, our metric gives better correlation in all language pairs. We performed statistical significance test to validate results from multiple runs and found that results are statistically significant under the one-tailed paired t-test at the 99\% signiﬁcance level.

The results for WMT'17 are shown in Table \ref{table: system17}. Our metric Bi-Di-Ent performs better than BLEU for all language pairs except for lv-en and tr-en pair. The correlation scores are also comparable to the top performing evaluation metrics. It achieves best scores in de-en and ru-en language pair among all metrics. Overall, our metric has the second highest average correlation score of 0.956, highest being 0.960. 

Table \ref{table: examples} shows a few examples of human evaluated score and score given by our metric from WMT'17 data. We observe that translations which are scored higher by human judges are scored higher by our metric as well.  

We observe some candidate-reference pairs where entailment relationship exists, but only in one direction. Such are mostly the cases where one side is more general than the other. The overall score given by our metric for such translation is low since we are taking the product of scores from both sides. This need not be always favourable while evaluating machine translations, as we have observed from the manual evaluations. 

\section{Conclusion and future work}
Semantic similarity is a suitable measure for machine translation evaluation and it correlates better with human evaluation compared to traditional n-gram based metrics. An effective way to determine this similarity is by checking if two sentences are paraphrases of each other. Bi-directional entailment can be used to determine paraphrasing. For translations to the English language we showed that fine-tuning sentence embedding model on MNLI corpus gives competitive results. Our approach can be further improved by incorporating external knowledge to handle the cases where only one-sided entailment is present because of generalization. Our work can be extended for evaluation of translations to other languages by using Cross-Lingual NLI (XNLI)\cite{conneau2018xnli} corpus. Bi-directional entailment can be applied to other NLP tasks also, like paraphrase detection.

\bibliography{emnlp-ijcnlp-2019}
\bibliographystyle{acl_natbib}

\end{document}